\documentclass[letterpaper, 10pt, conference]{ieeeconf}
\pdfoutput=1
\IEEEoverridecommandlockouts
\makeatletter
\let\NAT@parse\undefined
\makeatother

\usepackage{ifpdf}
\usepackage{amsmath,amssymb,amsfonts}
\usepackage{algorithmic}
\usepackage{graphicx}
\usepackage{textcomp}
\usepackage{xcolor}
\usepackage{mathrsfs}
\usepackage[numbers,sort&compress]{natbib} 
\usepackage{svg}
\usepackage{float}
\usepackage{csvsimple}
\usepackage[linesnumbered,ruled]{algorithm2e}
\usepackage{numprint}
\usepackage{balance}
\usepackage{booktabs} 
\usepackage{comment}

\usepackage[
hidelinks,
colorlinks,
linkcolor=black,
citecolor=black,
filecolor=red,
urlcolor=blue
]{hyperref}

\usepackage[capitalise]{cleveref}
\usepackage{caption}
\def\BibTeX{{\rm B\kern-.05em{\sc i\kern-.025em b}\kern-.08em
    T\kern-.1667em\lower.7ex\hbox{E}\kern-.125emX}}

\DeclareMathOperator*{\argmax}{arg\,max}

\newcommand{\actors}{\mathcal{A}}
\newcommand{\controlspace}{\mathcal{U}}
\newcommand{\numactors}{N^\mathrm{a}}

\newcommand{\robots}{\mathcal{R}}
\newcommand{\numrobots}{N^\mathrm{r}}

\newcommand{\faces}{\mathcal{F}}
\newcommand{\numfaces}{N^\mathrm{f}}

\newcommand{\states}{\mathcal{X}}

\newcommand{\objective}{\mathcal{J}}
\newcommand{\stationaryreward}{R_\mathrm{s}}
\newcommand{\viewreward}{R_\mathrm{v}}

\newcommand{\collisionrate}{r_\mathrm{c}}

\newcommand{\robotactions}{\mathscr{U}}
\newcommand{\matroid}{\mathscr{I}}
\newcommand{\ground}{\Omega}

\begin{document}
\graphicspath{{figures/}}

\title{\huge Greedy Perspectives: Multi-Drone View Planning for Collaborative
Perception in Cluttered Environments}
\author{%
Krishna Suresh$^{1}$, Aditya Rauniyar$^{2}$, Micah Corah$^{3}$, and Sebastian Scherer$^{2}$%
\thanks{$^{1}$K. Suresh is with Olin College of Engineering, Needham, MA, USA {\tt\small ksuresh@olin.edu}}%
\thanks{$^{2}$A. Rauniyar, and S. Scherer are with the Robotics Institute, School of Computer Science at Carnegie Mellon University, Pittsburgh, PA, USA
        {\tt\small \{rauniyar, basti\}@cmu.edu}}%
\thanks{$^{3}$M. Corah is with the Department of Computer Science at
  the Colorado School of Mines, Golden, CO, USA
  {\tt\small micah.corah@mines.edu}
}%
}

\maketitle


\begin{abstract}
Deployment of teams of aerial robots could enable large-scale filming of dynamic
groups of people (actors) in complex environments for applications in
areas such as team sports and cinematography.
Toward this end, methods for submodular maximization via sequential greedy
planning can enable scalable optimization of camera views across teams of
robots but face challenges with efficient coordination in cluttered
environments.
Obstacles can produce occlusions and increase chances of inter-robot collision
which can violate requirements for near-optimality guarantees.
To coordinate teams of aerial robots in filming groups of people in dense
environments, a more general view-planning approach is required.
We explore how collision and occlusion impact performance in filming
applications through the development of a multi-robot multi-actor view planner
with an occlusion-aware objective for filming groups of people and compare with
a formation planner and a greedy planner that ignores inter-robot collisions.
We evaluate our approach based on five test environments and complex
multi-actor behaviors.
Compared with a formation planner, our sequential planner generates 14\% greater
view reward for filming the actors in three scenarios and comparable performance
to formation planning on two others.
We also observe near identical view rewards for sequential planning both with
and without inter-robot collision constraints which indicates that robots are
able to avoid collisions without impairing performance in the perception task.
Overall, we demonstrate effective coordination of teams of aerial robots in
environments cluttered with obstacles that may cause collisions or occlusions
and for filming groups that may split, merge, or spread apart.
Our implementation and the data used to produce results for this paper are
available via the companion website:
\url{https://greedyperspectives.github.io/}
\end{abstract}

\section{Introduction}
The capture of significant events via photos and video has become universal, and Unmanned aerial vehicles (UAVs) extend the capabilities of cameras by allowing for view placement in otherwise hard-to-reach places and tracking intricate trajectories.
Multiple aerial cameras can be used to not only view an actor from multiple
angles simultaneously but perform higher functions such as
localization and
tracking~\citep{corah_scalable_2021,cai2023energy,schlotfeldt2021tro,hung2022tcns,zhao2019aamas},
environment exploration and mapping~\citep{corah2019auro,schlotfeldt2021tro},
cinematic filming~\citep{bucker_you_2021,alcantara_autonomous_2020,pueyo2024tro,nageli2017ral},
and outdoor human pose reconstruction~\citep{ho_3d_2021,saini2019markerless}.
These applications rely on effective collaboration between groups of UAVs
whereas manual control may result in poor shot selection and view duplication
while requiring many coordinated operators. Therefore, autonomous coordination
of UAV teams may be necessary for tasks such as multi-robot filming or
reconstruction.
However, directly maximizing domain-specific metrics, such as reconstruction
accuracy, can be difficult to perform online---this motivates development of
proxy objectives that quantify coverage and detail for multiple views.
For example, \citet{bucker_you_2021} demonstrate cinematic filming through a
joint objective combining collision and occlusion avoidance, shot diversity, and
artistic principles in filming a \emph{single actor}.
We are interested in similar settings but where robots collaborate to obtain
diverse views of \emph{multiple actors}---in a cluttered environment, with
occlusions, where robots may observe multiple actors at once.

\begin{figure}
\centerline{\includegraphics[width=\linewidth]{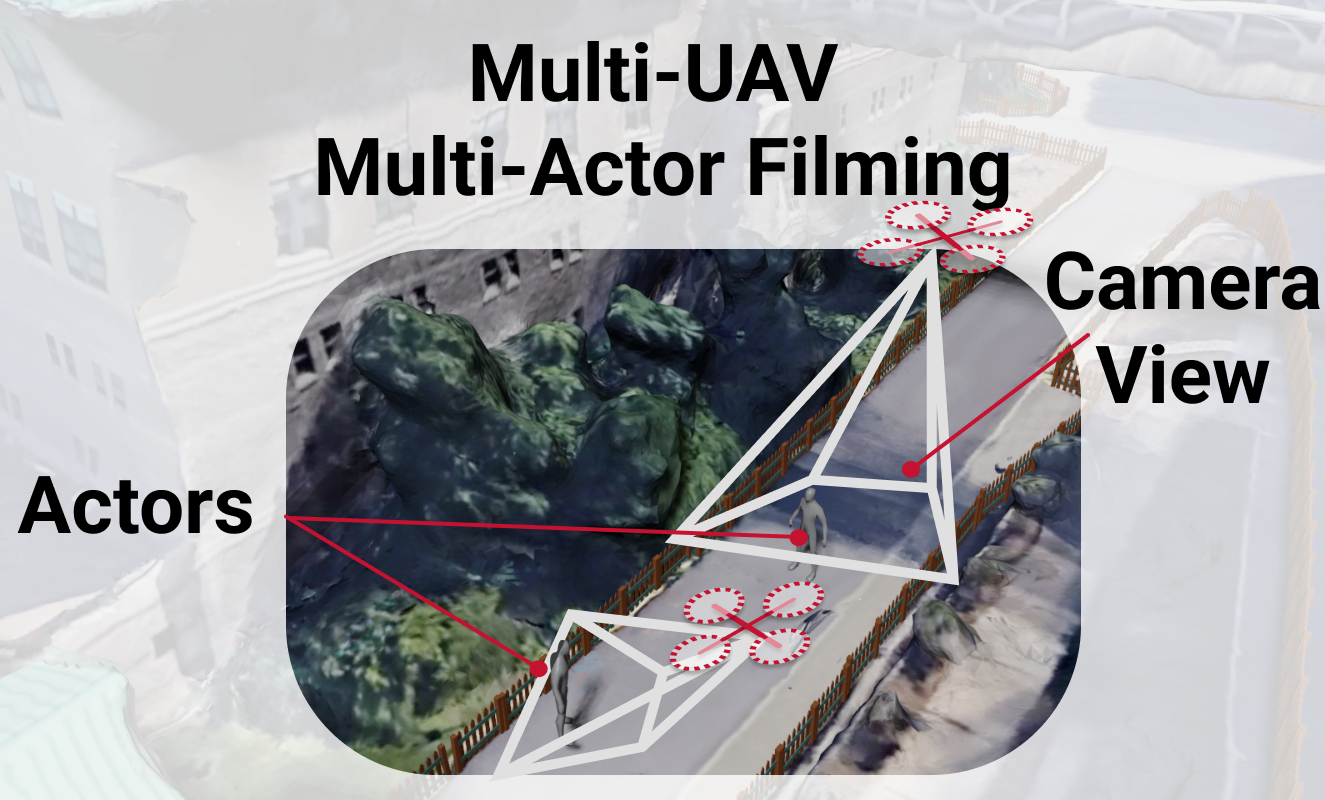}}
\caption{\textbf{Multi-Actor View Planning Scenario:} Known actor and environment geometries as well as actor trajectories and robot start locations are input into the view-planning system.
The planner aims to maximize coverage-like view reward for all actors for the
duration of the planning horizon. Mesh of CMU campus from \cite{cao2022icra}.}
\label{fig:problem}
\end{figure}

While defining an objective can be difficult, planning for multi-robot aerial
systems also presents a challenge: the vast joint state space,
non-convex environment, and non-linear view rewards make optimal planning
intractable.
Many applications exploit problem-specific structures to reduce the problem
complexity such
as by optimizing an actor-centric formation~\citep{ho_3d_2021,tallamraju2019ral,hung2022tcns}
or by altering the search procedure to generate single-robot trajectories
sequentially~\citep{bucker_you_2021,corah2019auro,corah_scalable_2021,cai2023energy,schlotfeldt2021tro}.
In this work, we apply a planning approach much like~\citet{bucker_you_2021},
and develop a system design and view rewards that enable application to a
multi-actor setting.

\begin{figure*}[h]
\centerline{\includegraphics[width=\textwidth]{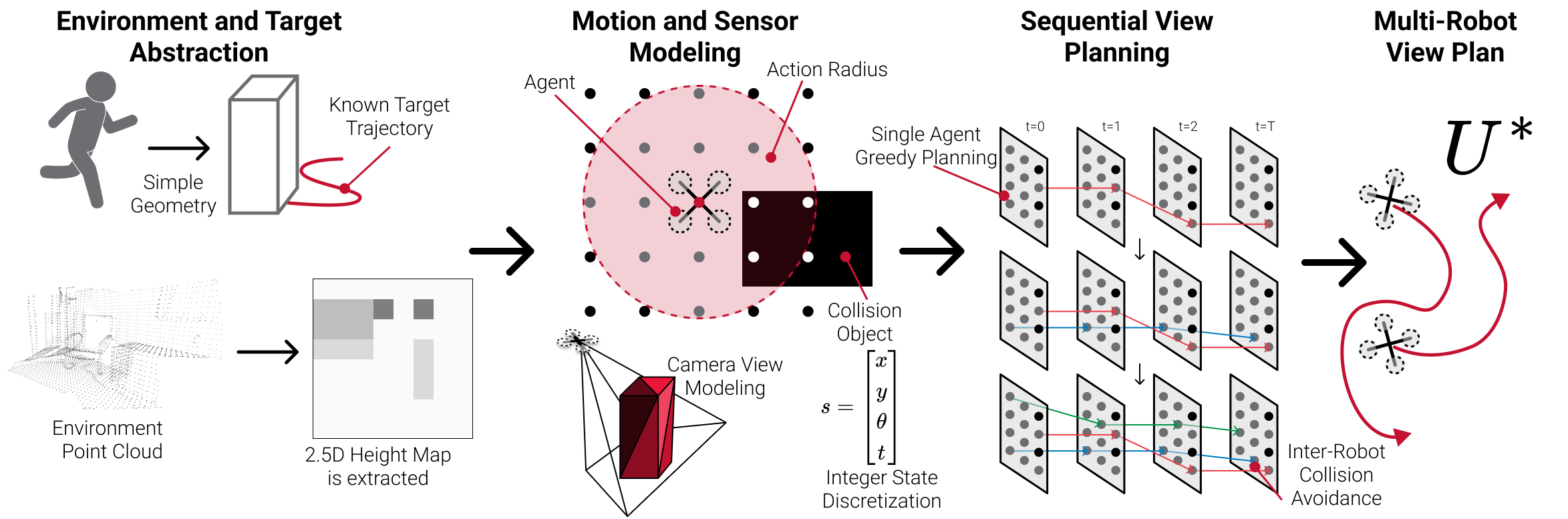}}
\caption{\textbf{View Planning System Overview} The multi-actor scenario is translated to an internal planner representation. The Markov Decision Process with DAG structure encodes collision constraints and view rewards. The Multi-Robot View Plan is produced through sequential greedy planning.}
\label{fig:system}
\end{figure*}

\textbf{Problem:} The dynamic multi-actor view planning problem consists of
generating sequences of camera views over a fixed planning horizon to maximize a
collective view reward that is a function of pixel densities over the surfaces
of the actors.
The primary assumptions for our approach are:
\emph{known static environment},
\emph{known actor trajectories} (e.g. scripted scenarios),
and \emph{known robot start state}.
An illustration of the problem setup is depicted in \cref{fig:problem}.
These assumptions are fairly strict for the purpose of evaluating the planning
approach.
In practice, our approach could be applied in a receding-horizon setting based
on scripted or predicted actor trajectories~\cite{bucker_you_2021}.



\textbf{Contributions:} The main contributions of this work are summarized as:
\begin{itemize}
    \item An occlusion-aware objective based on rendered camera views for
      filming groups of people
    \item Implementation of a collision-aware multi-robot, multi-actor view
      planner
    \item Evaluation of the view planner in scenarios with
      challenging obstacles, occlusions, and group behaviors
\end{itemize}
Our contributions build on prior work developing perception objectives based on
pixel densities by \citet{jiang_onboard_2023} and work by
\citet{hughes2024cdc} developing objectives for submodular multi-robot
settings.
This work presents a variant of such objectives that is occlusion-aware,
and we present results for scenarios with a wide variety of obstacles and
occlusions.
These contributions toward objective design also enable us employ a planning
approach similar to \citet{bucker_you_2021} but for filming multiple actors.

\section{Related Works}
\textbf{Aerial Filming:} Aerial perception systems have grown to widespread use
through their success in providing low-cost filming of conventionally
challenging unscripted scenes.
Consumer and commercial systems such as the Skydio S2+
\cite{noauthor_drone_nodate} demonstrate single-drone filming capabilities and
are starting to incorporate collaborative multi-drone behaviors for mapping.
Research developing autonomous aerial filming systems for cinematography has
also focused on developing systems that can perform parametrized actions
such as tracking a subject, rule of thirds framing, or dolly zoom
by methods such as model predictive control to optimize motion and camera
controls~\cite{pueyo2024tro,nageli2017ral} and by scheduling
multi-robot
missions~\citep{alcantara2020ieee,mademlis2023,caraballo2020iros,ray2021iros}.
Another body of work focuses on learning artistic principles rather than
implementing them by hand~\citep{pueyo2023iros,bonatti2020jfr}.
The focus of these works on mission planning and learning
artistic principles is complementary to our approach which develops a capability
for filming complex group behaviors with occlusions.

A few works address this challenge of filming individual or group
behavior from multiple
perspectives~\citep{xu2022iros,bucker_you_2021}.
\citet{xu2022iros} present an actor-centered controller based on Voronoi
coverage over a hemisphere, and
\citet{bucker_you_2021} describe an approach where robots plan trajectories
based on a spherical discretization centered on the actor.
A limitation of these approaches is that assigning robots to actors does not
exploit the robots' capacity to observe multiple actors at once.

\textbf{Motion capture and dynamic scenes:}
Aerial motion
capture~\citep{saini2019markerless,ho_3d_2021,aircaprl,tallamraju2019ral} such
as to reconstruct the motion of a moving person is closely related to filming.
So far, these systems consist of a groups of robots that observe a single moving
subject with various degrees of awareness of obstacles or occlusions.
However, all specify either an actor-centric policy~\citep{aircaprl} or a
formation~\citep{saini2019markerless,ho_3d_2021,tallamraju2019ral}.
Although actor-centric planning can reduce the search space to orienting the
formation versus planning for robots individually, this choice limits relevance
to multi-actor settings.
The view planning approach we present could enable robot teams to reconstruct
motions of complex group behaviors in environments with varied obstacles and
occlusions.

\textbf{Reconstruction of static scenes:}
Our view planning approach also bears similarity to methods for reconstruction
of static scenes~\citep{jiang_onboard_2023,delmerico2018auro,roberts2017iccv}.
Like these works, our approach emphasizes design of a view reward and optimizing
paths to maximize that reward.
In particular, we draw on the approach by~\citet{jiang_onboard_2023} which
reasons about pixel densities.

\textbf{Target tracking and localization:}
The filming scenario we study is similar in composition to target localization
and tracking problems that focus on estimating position or motion of one or
more
targets~\citep{corah_scalable_2021,cai2023energy,schlotfeldt2021tro,hung2022tcns,zhao2019aamas}.
While these works sometimes reason about field-of-view and
occlusions~\citep{hung2022tcns}
this reasoning is often limited and secondary to the tracking task---a task that
is often formulated in terms of minimizing uncertainty in target state via an
information-theoretic
objective~\citep{corah_scalable_2021,cai2023energy,schlotfeldt2021tro,zhao2019aamas}.
Target tracking may also be thought of as representing a
subsystem of a hypothetical filming system that relaxes our requirement for
known trajectories---from this perspective, we observe that several works in
this area apply sequential and auction-based methods for coordination that are
generally compatible with our
approach~\citep{corah_scalable_2021,cai2023energy,schlotfeldt2021tro,zhao2019aamas}.

\textbf{Sequential and Submodular Multi-Robot Planning:}
Typically multi-robot perception planning and information gathering problems,
cannot tractably be solved optimally due to their combinatorial nature, but
greedy methods for submodular optimization can often promise information gain or perception
quality no worse than half of optimal in polynomial
time~\citep{nemhauser1978,fisher1978}.
Submodular optimization and sequential greedy planning has been applied
extensively to such multi-robot coordination
problems~\citep{singh_efficient_2009,corah2019auro,corah_scalable_2021,cai2023energy,schlotfeldt2021tro,lauri_multi-sensor_2020,mccammon2021jfr,bucker_you_2021}.
However, questions of occlusions and camera views have been explored primarily
in the setting of mapping and
exploration~\citep{corah2019auro,schlotfeldt2021tro}.
\citet{lauri_multi-sensor_2020} present an exception in which eye-in-hand
cameras inspect and reconstruct static scenes.
Unlike exploration and mapping, applications involving filming moving actors
can force persistent interaction between robots over the duration of a scenario
or planning horizon, and our early work on this topic indicates that
sequential planning is important for effective cooperation in settings
involving observing moving subjects~\cite{corah_performance_2022}.

\section{Preliminaries}
We will begin with some background regarding submodular and greedy planning:

\subsection{Submodularity and Monotonicity}
The view reward we employ satisfies monotonicity properties that are useful in
developing our approach to planning and coordination.
Informally, submodularity expresses the principle of diminishing returns and
monotonicity requiring functions to be always increasing.
Given a set of actions $\ground$, a set
function
$g$
maps subsets of actions (robots'
plans) to a real value (the view reward).
A set function is monotonic if adding elements to a set
does not decrease its value; that is
for any $A \subseteq B \subseteq \ground$ then $g(A) \leq g(B)$.
A set function is submodular if marginal gains decrease monotonically;
specifically, given $A \subseteq B \subseteq \ground$ and $C
\subseteq \ground \setminus B$, then
$g(A \cup C) - g(A) \geq g(B \cup C) - g(B)$.
Objectives related to perception
planning~\citep{corah2021icra,lauri_multi-sensor_2020} and information
gathering~\citep{singh_efficient_2009} are often submodular, and this
corresponds to how marginal view rewards may diminish given by repeated views of
the same actor from the same perspective.

\subsection{Partition Matroids}

A partition matroid can be used to represent product-spaces of actions
or trajectories that arise in multi-robot planning problems~\cite[Sec.
39.4]{schrijver_combinatorial_2003}.
Consider a view planning problem involving
a team of robots
$\robots=\{1,\ldots,\numrobots\}$
where each robot $i \in \robots$
 has access to a set of actions
$\robotactions_i$.
These actions can take many
forms such as assignments, trajectories, or paths.
The set of all actions for a
robot is the ground set $\ground=\bigcup_{i \in \robots}\robotactions_i$.
Each robot is assigned one action from its corresponding set
$\robotactions_i$. 
If there are no collisions between robots the set of valid and partial
assignments forms a \textit{partition matroid}:
$\matroid=
\{X\subseteq\ground \mid 1 \geq |X \cap \robotactions_i|\ \forall i \in
\robots\}$.
To satisfy this structure each robots' actions must be interchangeable to satisfy
the \emph{exchange property} of a matroid.
Inter-robot collisions violate this
property because swapping actions can cause conflicts with other robots.

\section{Problem Formulation}

We aim to coordinate a team of UAVs to maximize coverage-like view rewards
for observing a group of actors moving through an obstacle-dense environment.
Consider a set of actors $\actors=\{1,\ldots,\numactors\}$ each with a set of
faces $\faces_a = \{1,\ldots,\numfaces_{a}\}$ where $a \in \actors$ and a set of
robots $\robots=\{1,\ldots,\numrobots\}$.
Each robot $i \in \robots$ can execute a control action
$u_{i,t} \in \controlspace_i \subseteq SE(2)$
at time $t \in \{0,\ldots,T\}$.
The robots go on to select a finite-horizon sequence of viable control
actions that form their plans.
Additionally, robots have associated states $x_{i,t} \in \states$ which is
a subset of $SE(2)$.
Sequences of states form the robots' trajectories
$\xi_i = [x_{i,0},\ldots,x_{i,T}]$%
---we will occasionally index trajectories to obtain $\xi_{i,t}=x_{i,t}$.
Each trajectory, once fixed, produces non-collision constraints for all other
robots.
Given the trajectories of all actors in $SE(3)$, start states $x_{i,0}$, and
environment geometry, we aim to find joint collision-free control sequences
$U^\ast = \bigcup_{i \in \robots} [u_{i,0},\ldots,u_{i,T}]$ that
maximize our objective and fit our motion model.

\subsection{Motion Model}
\label{sec:motionModel}
State transitions for each robot are specified by the following motion model
\begin{align}
  x_{i,t+1}=f_i(x_{i,t},u_{i,t})
  \label{eq:dynamics}
\end{align}
where $f_i$ is defined to only allow collision-free motions to positions and
orientations within a constant velocity constraint.
Given the time step duration, maximum translational and rotational velocities are converted to bounds on rotation and Euclidean distance as illustrated by \cref{fig:system}.

\subsection{Non-Collision Constraints}
\label{sec:collision_constraints}
Robots are considered in collision with the environment when the discretized
state location is occupied by an element of the environment map that exceeds the
robot's height.
Similarly, a pair of robots is in collision when both occupy the same
discretized cell at the same time.
We implicitly assume a conservative model of the environment (the height-map and
discretization of the state space) to ensure safety of states that satisfy
the obstacle and inter-robot collision constraints.

\subsection{Camera and View Reward Model}
We use a coverage-like reward based on pixel densities as a proxy for
effectively observing an actor.
Inspired by \cite{jiang_onboard_2023},
we compute rewards based on cumulative pixel densities ($\frac{px}{m^2}$) for
observations of faces from polyhedral representations of each actor $j$.
We define a function $\texttt{pixels}(x_{i,t}, t, j, f) \to \mathbb{R}$ which
returns the pixel density for actor-$j$'s face $f$ when observed from a
robot's state at time $t$.

In order to encourage robots to assume views that uniformly cover the actors and
their corresponding faces, we apply a square root to introduce diminishing
returns on increasing pixel densities from multiple
views~\citep{hughes2024cdc}.
Finally, given robot trajectories according to the dynamics \eqref{eq:dynamics}
and selected control inputs, the robots obtain the following view reward for the
given face and time:
\begin{align}
  \viewreward(t, j, f) =
  \sqrt{\sum\nolimits_{i\in \robots} \texttt{pixels}(x_{i,t}, t, j, f)}.
  \label{eq:view_reward}
\end{align}
The formal statement of monotonicity and submodularity properties for rewards of
this form and the relationship to coverage are the subject
of~\citep{hughes2024cdc}.
Intuitively, \eqref{eq:view_reward} obtains these desirable properties
because the square-root is one of many real functions that increases
monotonically but at ever-decreasing rates.\footnote{%
  In fact, any other real function with similar monotonicity properties other
  than the square-root would
  satisfy the requirements of submodularity and monotonicity.
  However, we will not focus on the choice amongst such functions in this work.
}
If not constrained through the selection of camera views, the submodularity
of our objective based on \eqref{eq:view_reward} would ensure that rewards
would be maximized by distributing all pixels approximately uniformly over the
faces of the actor models.
Likewise, submodularity due to the square-root encourages robots to distribute
their views evenly across the actors and their surfaces.
By contrast, summing pixel densities without the square-root
would assign the same reward for distributing all pixels on one face or for
distributing pixels uniformly.
Our approach also allows for more variation in rewards than for simply
thresholding on range or pixel density.

\subsection{Objective}
In addition to maximizing the view reward, we add a reward for stationary behavior
$R_s(u_{i,t})$ to reduce unnecessary movement whereas
\begin{align}
  R_s(u) =
  \begin{cases}
    \epsilon & \text{if } u \text{ is stationary} \\
    0 & \text{otherwise}.
  \end{cases}
  \label{eq:stationary}
\end{align}
So, robot $i\in\robots$ obtains
$\sum_{t\in \{0,\ldots,T\}} R_s(u_{i,t})$ reward for time-steps it remains
stationary.
The joint objective is then as follows:
\begin{align}
  \begin{split}
    &\objective(X_{\texttt{init}}, U) =
    \\
    &\quad
    \sum_{t\in \{0,\ldots,T\}}
    \bigg(
      \sum_{i \in \robots}
      \stationaryreward(u_{i,t})
      +
      \sum_{j \in \actors} \sum_{f \in \faces_j}
      \viewreward(t,j,f))
    \bigg)
  \end{split}
  \label{eq:reward}
\end{align}
where $X_{\texttt{init}}=[x_{0},\ldots,x_{\numrobots}]$ is an array of
initial robot states and $U$ represents the robots' sequences of control
actions.
Since we aim to find the control sequence that maximizes this objective, our
optimal control sequence can be defined as:
\begin{align}
  U^\ast = \argmax_{U} \objective(X_{\texttt{init}}, U)
  \label{eq:optimization_problem}
\end{align}

\section{Multi-Robot Multi-Actor View Planning}

We now present our multi-drone view planning approach.
The planner aims not only to produce sufficient coverage over the actors but
also to exploit problem structure to efficiently find single-robot trajectories
by greedy planning.

\begin{figure}[h]
  \vspace{1ex}
\centerline{\includegraphics[width=\linewidth]{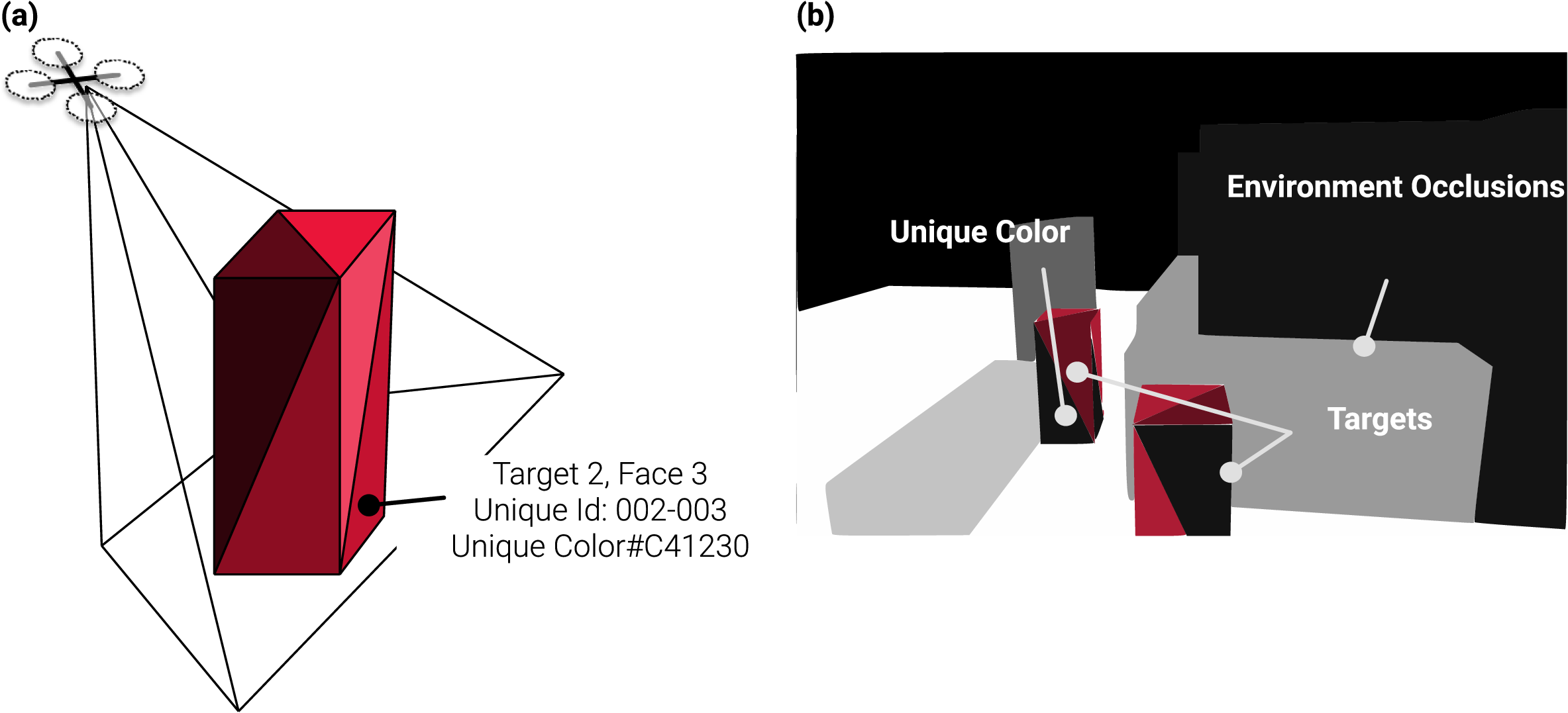}}
\caption{\textbf{Actor Coverage} \textbf{(a)} UAV camera model frustum observing a simplified actor geometry. Actor faces are colored slightly differently based on a face identification system to allow for pixel density computation.  \textbf{(b)} Example camera output from OpenGL internal rendering system.}
\label{fig:coverage}
\end{figure}

\begin{figure*}[h]
  \centerline{\includegraphics[width=\linewidth]{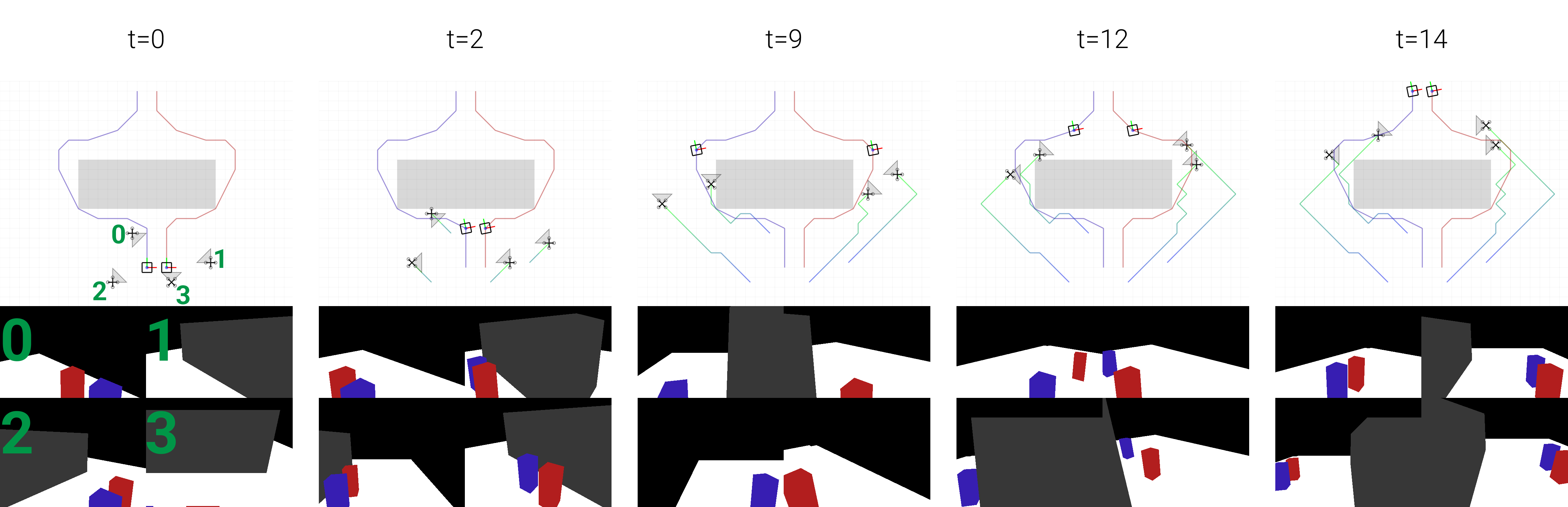}}
  \caption{Example robot views and joint trajectories from sequential plan in $\texttt{Split}$ test case. Robot first-person views at each time step display viewing of the actors over the planning horizon.}
  \label{fig:traj}
\end{figure*}
\subsection{Evaluation of View Reward}
To produce an occlusion-aware implementation of our view reward
\eqref{eq:view_reward} we compute $\texttt{pixels}$ by implementing an OpenGL
rasterization renderer based on a 2.5D height map of the environment and
simplified actor geometry---we use polygonal cylinders.
We then use a perspective camera based on specified camera intrinsics to
capture an occlusion-aware representation of the scene from a given robot state.
The system renders the environment via the GPU with a geometry shader to draw
the height-map.
To determine how many pixels the cameras observe for each face, we render the
faces with unique colors associated with the actor and face IDs.
Counting pixels of each color and dividing by the areas of the faces yields
the corresponding pixel densities.
\Cref{fig:coverage} illustrates this process and provides an example of a
rendered view.

\subsection{Single-Robot Planning}
\label{sec:single_robot}
With the robot state in $SE(2)$ we aim to represent the single-robot planning problem as
a Markov Decision Process (MDP) which has an underlying
Directed Acyclic Graph (DAG) structure.
We use the AI-ToolBox library to represent and solve the MDP~\cite{noauthor_svalorzenai-toolbox_nodate}.
The MDP state $s$ is represented as an integer vector:
  \begin{align*}
    s &= \begin{bmatrix}
           x &
           y &
           \theta &
           t
         \end{bmatrix}
  \end{align*}
Each MDP action $a$ is in the same discrete space, encoding the next state and incrementing the time by 1.
This forces the MDP structure to be directed since states can never go back in time.
The MDP is constructed with a transition matrix associating each $(s,a,s^{\prime})$ tuple with a transition probability and a reward matrix associating each $(s,a)$ pair with a reward according to \eqref{eq:reward} with knowledge other robots' actions (to be introduced in \cref{sec:sequential}). We perform a breadth-first search over the state space by branching on feasible actions to populate the transition and reward matrices. As depicted in \cref{fig:system}, the set of available actions is pruned based on environmental and inter-robot collisions. This directed MDP can be solved with one backward pass of value iteration to find the optimal greedy policy,\footnote{Our current implementation converges in 5 passes without exploiting this structure.} similar to the approach by~\citet{bucker_you_2021}.
Following this policy from the initial state produces the optimal single-robot control sequence.

\subsection{Sequential Planning}
\label{sec:sequential}
Now, we are able to generate the joint view plans for the multi-robot team.
We do so by sequentially planning greedy single-robot trajectories in an arbitrary order as is common
for methods based on submodular
optimization~\citep{singh_efficient_2009,cai2023energy,bucker_you_2021,schlotfeldt2021tro,corah_scalable_2021,corah2019auro,lauri_multi-sensor_2020,mccammon2021jfr}.
With some abuse of notation, each robot maximizes the objective for itself with
access to prior decisions in the sequence:
\begin{align}
  U_i = \argmax_{U} \objective(X_{\texttt{init},1:i}, U_{1:i-1} \cup U)
  \label{eq:sequential}
\end{align}
This forms a series of single-robot sub-problems that we solve with the value
iteration planner (Sec.~\ref{sec:single_robot}).
Through the course of this process, we accumulate pixel densities per each face
to evaluate the view reward \eqref{eq:view_reward} and filter out states
that would produce collisions with other robots (Sec.~\ref{sec:collision_constraints}).

\subsubsection{Suboptimality guarantees and inter-robot collisions}
If we ignore inter-robot collisions, \eqref{eq:optimization_problem} has the
form of a submodular maximization problem with a partition matroid constraint.
Thanks to the famous result by \citet{fisher1978},
sequential greedy planning via \eqref{eq:sequential} is guaranteed to produce
solutions to \eqref{eq:optimization_problem} with objective values no worse than
50\% of optimal.
However, inter-robot collisions violate the form of a partition
matroid~\cite{corah2019auro} so solutions that incorporate inter-robot
non-collision constraints will not satisfy this guarantee.
However, if the operating environment is not congested with robots, these
non-collision constraints may not significantly influence the view rewards in
practice.
Our simulation results in \cref{sec:results_planners} support this
conclusion that inter-robot constraints have negligible impacts
on solution quality while \emph{averting the serious consequences of collision}.

\subsection{Time Scaling Analysis}
\label{sec:asymptotic_analysis}

This section seeks to clarify the computational cost of \cref{alg:seq}.
After instantiating the MDP, value iteration for a single robot runs (ideally)
with a single backwards pass over the reachable states.
For planar motion the number of reachable states at step $t$ is $O(t^2)$,
and the total number of states over a horizon of $T$ steps is $O(T^3)$.
Inter-robot collision checking involves computation for each prior robot at
every state, and for a single robot requires
$O(T^3\numrobots)$ time.
The total time for the entire sequential planning process is then
$O(T^3{\numrobots}^2)$.
This addresses number of robots but not necessarily increasing problem scale in
terms of the number of actors or environment complexity.
Incorporating larger environments or more actors introduces more nuance. For
example, evaluating the objective \eqref{eq:reward} is at least
proportional to $\numactors$ but may be larger depending on the cost of
rendering scenes with different numbers of actors.
So, we can also say that the computation time scales as
$\Omega(T^3{\numrobots}^2\numactors)$.

\subsection{Considerations for application to real systems}

Now, let us discuss how the proposed approach could be adapted for
implementation on physical robots.
First, our approach relies on known or predicted actor trajectories.
This is reasonable for a scripted sequence such as when filming a movie---in
this case, offline planning may also be reasonable.
However, to account for uncertainty in actor or robot motions, systems should
employ receding-horizon planning; though practical application would require
substantial improvement in planning time.
Then, for outdoor operation, GPS is often sufficient for
localizing robots~\citep{ho_3d_2021,bonatti2020jfr,saini2019markerless},
particularly with RTK systems.
Although prior works have implemented visual tracking for a single
actor~\citep{bonatti2019iros},
additional instrumentation on the actors (such as additional GPS units) may be
preferable for multi-actor settings.
Then, while long-horizon prediction may be challenging, a Kalman filter can
provide predictions over a short horizon~\citep{bonatti2019iros} based on
velocity and orientation.
Regarding the map, if robots operate frequently in the same location
(like a sports arena), a pre-existing map along with local collision avoidance
may be appropriate.

 \begin{algorithm}[h]
 \DontPrintSemicolon
 \caption{Sequential Greedy View Planning
 }\label{alg:seq}
Initialize $U_{\text{seq}} \gets \{\}$\;
Initialize collisionMap $\gets \{\}$\;
\ForEach{$i$ in $\robots$}{
    $S_i \gets$ DiscretizeStateSpace(envHeightMap)\;
    $A_i \gets$ DiscretizeActionSpace(robotMaxMotion)\;
    MDP $\gets$ BreadthFirstSearch($x_{i,0}$, $S_i$, $A_i$)\;
    \tcp{In BFS, $\viewreward$ is computed at each explored state. Branching through $\texttt{availableActions}$ removes actions that lead to collision states.}
    $\pi_i \gets$ ValueIteration(MDP)\;
    $\{u_{i,0},\ldots, u_{i,T}\}\gets$ ExtractTrajectory($\pi_i$)\;
    Append $\{u_{i,0},\ldots, u_{i,T}\}$ to $U_{\text{seq}}$\;
    $\xi_i \gets$ applyActions($x_{i,0}$,$\{u_{i,0},\ldots, u_{i,T}\}$)\;
    addCollisions($\xi_i$)\;
}
return $U_{\text{seq}}$\;
 \end{algorithm}

\section{Experiments}
\label{sec:experiments}
We evaluate the performance of the sequential view planner in five test scenarios that aim to demonstrate view planning under a variety of conditions, and we compare to a formation planning baseline.

\subsection{Formation Planning}
\label{sec:formation_planner}

\begin{figure}
\centerline{\includegraphics[width=\linewidth]{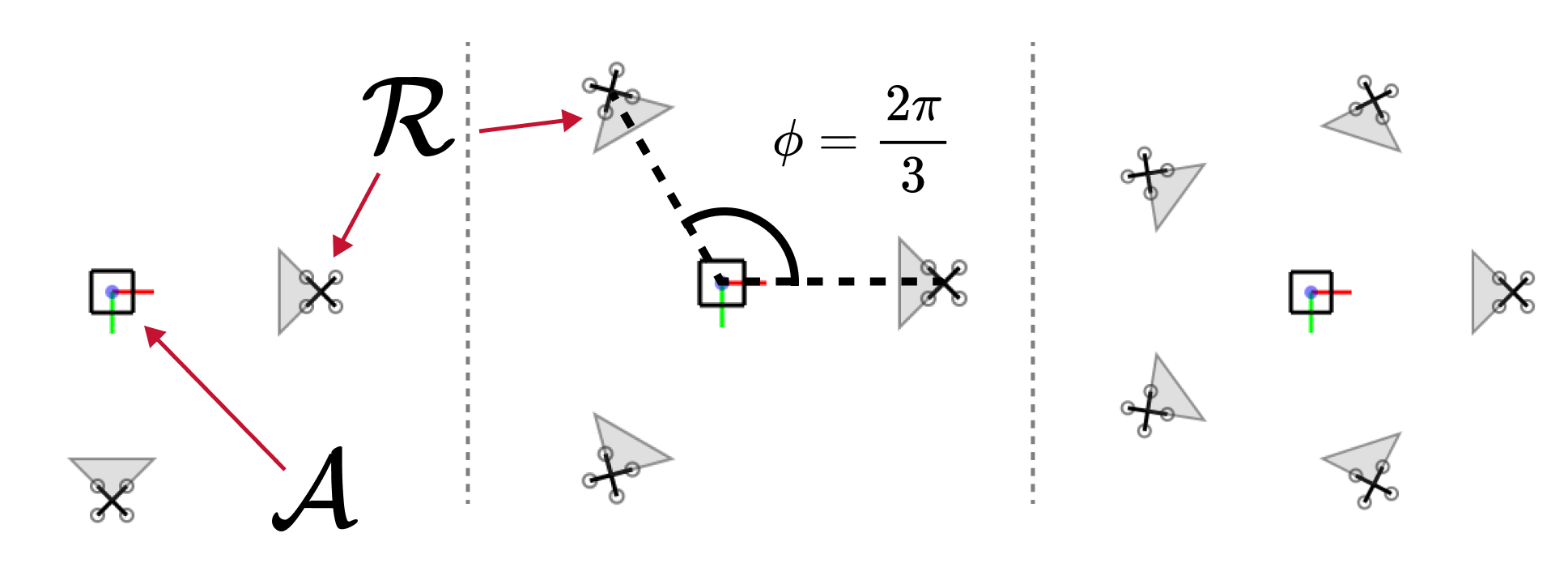}}
\caption{Multi-robot formations for $\numrobots=2,3,5$}
\label{fig:formations}
\end{figure}
\begin{figure*}[h]
  \centerline{\includegraphics[width=\linewidth]{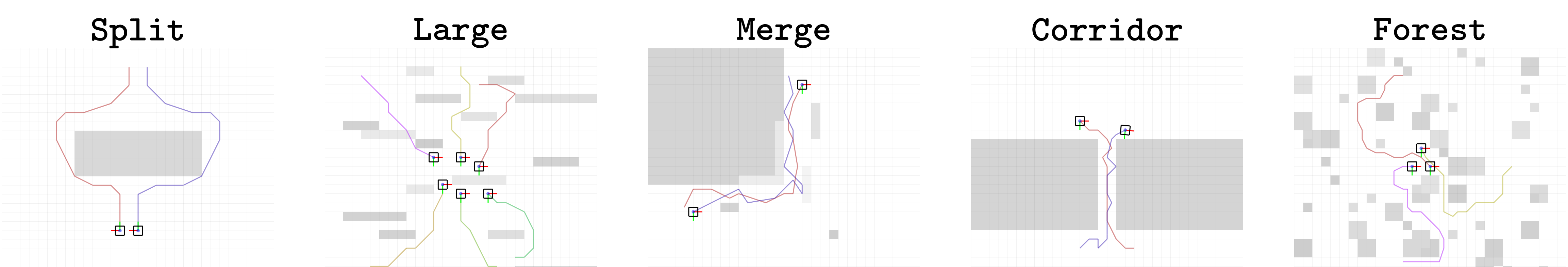}}
  \caption{Scenarios to evaluate specific aspects of multi-actor view planning. Actors are illustrated as boxes with uniquely colored trajectories. The darkness of elements in the height map indicates their occupied height. $\texttt{Large}$ is the only test case with no collision obstacles as all elements are below the robot operating plane. }
  \label{fig:tests}
\end{figure*}

We compare our sequential view planner against an assignment and formation based
planner which we model off the multi-view formations applied
in \citep{tallamraju2019ral,ho_3d_2021} following analysis
by~\citet{bishop2010auto}.
We assign equal numbers of robots to each actor, and groups assume formations as
follows.
The formation has a constant radius around an actor and a separation
angle $\phi$;
for $\numrobots > 2$ then $\phi=\frac{2
\pi}{\numrobots}$  and
when $\numrobots=2$ then $\phi=\frac{\pi}{2}$
(see \cref{fig:formations}).
We directly orient the formation about the actor to maximize the view
reward $\viewreward$ (including all actors) at each time-step.
The formation planner has a fixed radius, ignores the motion model (Sec.~\ref{sec:motionModel}), and does
not consider environment and robot collisions; so view rewards for formation
planning are generally optimistic and require no-extra computation.

\subsection{Test Scenarios}

\begin{table}
    \centering
    \begin{filecontents*}{data/tests.csv}
test,robots,actors,timesteps,obstacles
Merge,4,2,17,Yes
Corridor,2,2,17,Yes
Forest,2$^{*}$,3,20,Yes
Large,18,6,10,No
Split,4,2,15,Yes
\end{filecontents*}
\csvreader[tabular = ccccc, table head = \bfseries Test Name & \bfseries \# Robot & \bfseries \# Actors& \bfseries Timesteps & \bfseries Env Collision \\\hline]{data/tests.csv}{}{
        \texttt{\csvcoli} & \csvcolii & \csvcoliii & \csvcoliv & \csvcolv
    }
    \caption{Test scenarios and parameters. The formation planner obtains an extra robot in the \texttt{Forest} scenario ``for free'' to match the number of actors}\label{table:scenarios}
\end{table}

Test scenarios are detailed in \cref{fig:tests} and \cref{table:scenarios}.
We use robots with camera intrinsic parameters of 2500px, 4000px, and 3000px
(focal length, image width, image height).
All drones are placed at 5 meters high with a camera tilt of 10 degrees from the horizon.
For each test scenario, we also specify 10 unique robot starting configurations to introduce further variation.
The scenarios are as follows:

$\texttt{Split}$: This test case is a simple group split and merge of 2 actors around an obstacle. A full view sequence from an example trajectory is displayed in \cref{fig:traj}.

$\texttt{Large}$: Focuses on scaling to larger teams and features 18 robots. Actors move through a series of short walls that produce occlusions but not collision constraints since they are below the navigation plane.

$\texttt{Merge}$: Contains actors moving around a corner in opposite directions. This test case investigates implicit actor assignment with actors being ``handed off'' at the corner.

$\texttt{Corridor}$: Tests robots moving through a narrow corridor. This focuses on the collision-aware aspect of the planner.

$\texttt{Forest}$:
This is a dense occlusion/collision environment.
For this scenario we limit sequential planning to two robots, fewer than the
number of actors (three).
This test aims to demonstrate the capacity to adapt to scenarios where assignments are not possible by evaluating if fewer robots can achieve similar or better coverage compared with the formation planner which requires 3 robots due to the minimum of 1 robot per formation.
For the purpose of evaluation, we treat both planners as featuring only two robots when we report \emph{per-robot} results.

\subsection{Sequential Planner Performance}
\label{sec:results_planners}

\begin{table}
  \centering
\begin{tabular}{lcclc}
  \toprule
  \bfseries Test & \bfseries Formation & \multicolumn{2}{c}{\bfseries Seq. w/o Inter-Robot} & \bfseries \emph{Sequential} \\
  \midrule
  \texttt{Split}    &     \textbf{1380} &   1352 $\pm$    34       & \textcolor{red}{$\collisionrate=0.4$} &    1351 $\pm$      35          \\
  \texttt{Large}    &     \textbf{1413} &   1390 $\pm$    27       & \textcolor{red}{$\collisionrate=10.7 $} &    1381 $\pm$      27          \\
  \texttt{Merge}    &     1149          &   \textbf{1275 $\pm$ 26} & \textcolor{red}{$\collisionrate=0.2 $} &    1274 $\pm$      28          \\
  \texttt{Corridor} &     1612          &   1808 $\pm$    86       & \textcolor{red}{$\collisionrate=0.8 $} &    \textbf{1812 $\pm$      85} \\
  \texttt{Forest}   &     2114          &   \textbf{2534 $\pm$ 73} & \textcolor{red}{$\collisionrate=0.2 $} &    2505 $\pm$ 116              \\
  \bottomrule
\end{tabular}

    \caption{Average and standard deviation of view reward ($\viewreward$) per
      robot for all test cases from 10 robot start configurations, comparing
      baselines to our approach (sequential).
      For sequential planning without inter-robot constraints, we also report the collision rate
      $\collisionrate$ (robots collided per trial).
    }\label{table:results}
\end{table}

\begin{figure}[h]
  \centerline{\includegraphics[width=\linewidth]{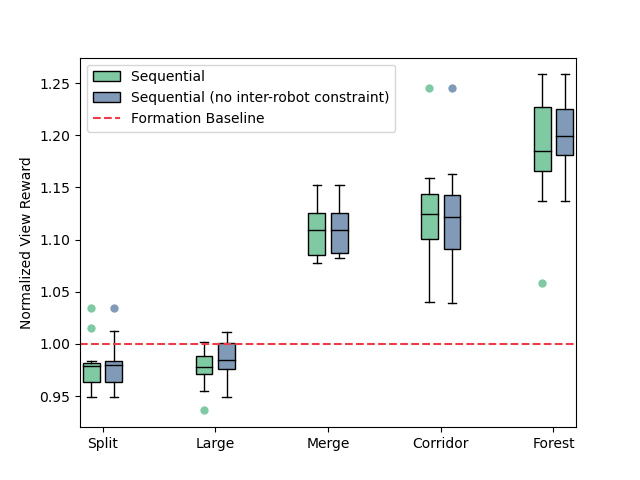}}
  \caption{Average view reward normalized by the baseline formation planner performance. 10 unique robot start configurations were specified for the sequential planners and are compared with the unique output from the formation planner.
  Both sequential planners outperform the formation planner baseline in the
  \texttt{Merge}, \texttt{Corridor}, and \texttt{Forest} scenarios, or else
  perform similarly.
  Sequential planners (with and without collision avoidance) also behave
  similarly in terms of view rewards: view rewards for the collision-aware
  planner are not impaired, though that planner no longer satisfies guarantees
  on solution quality.
  }
  \label{fig:results}
\end{figure}

\cref{fig:results} and \cref{table:results} summarize planner performance
across each of the scenarios in terms of the view reward for formation planning
and sequential planning both with and without inter-robot collision
constraints.
We observe that sequential planning\footnote{Referring to the collision-free
version, but both obtain similar performance.} outperforms formation planning
in three of five scenarios---by an average of 13.9\% in the $\texttt{Merge}$,
$\texttt{Corridor}$ and $\texttt{Forest}$ scenarios.
Notably, sequential planning also outperforms formation planning by 18\% in the
\texttt{Forest} scenario despite having one fewer robot.
We also observe that inter-robot non-collision constraints do not significantly
impair the performance of sequential planning.
In the $\texttt{Split}$ and $\texttt{Large}$ test cases, all planners perform
similarly.
This may be because these scenarios provide more favorable conditions for the
formation planner (and because the formation planner does respect starting
positions or robot dynamics).
\texttt{Split} provides ample space for formations of two robots to view the
actors, and
\texttt{Large} has shorter obstacles that produce occlusions but not collisions.

The sequential planner that ignores inter-robot collisions
guarantees solutions within half of optimal but may allow robots to collide.
Since both versions of the sequential planner obtain similar solution quality
(Fig.~\ref{fig:results}), we conclude inter-robot collision constraints do not
significantly impair performance in this setting.
We also report collision rates $\collisionrate$ in terms of
\emph{robots collided per trial}
for the sequential planner that ignores collisions \cref{table:results}.
In most scenarios, we observe less than one collision per trial except for
\texttt{Large} where ten robots (more than half) collide on average, but we
still do not see a significant difference in objective values.
The $\texttt{Corridor}$ scenario is also highly-constrained, and we see similar
objective values despite increased collisions.

\Cref{fig:traj} displays the joint view plans and internal view planner renderings for the $\texttt{Split}$ test case.
This figure illustrates the capacity of sequential planning to optimize views of one or more actors and to implicitly reconfigure or ``hand-off'' assignments over the course of a trial.
The robots all view both actors at $t=2$; the pairs split off and transition to each viewing a single actor by $t=9$; and they go back to jointly viewing actors by $t=14$.
This behavior is not manually specified and arises only from optimizing trajectories and views.

\subsection{Scaling number of robots}
\label{sec:scaling_results}
\begin{figure}[h]
    \centerline{\includegraphics[width=\linewidth]{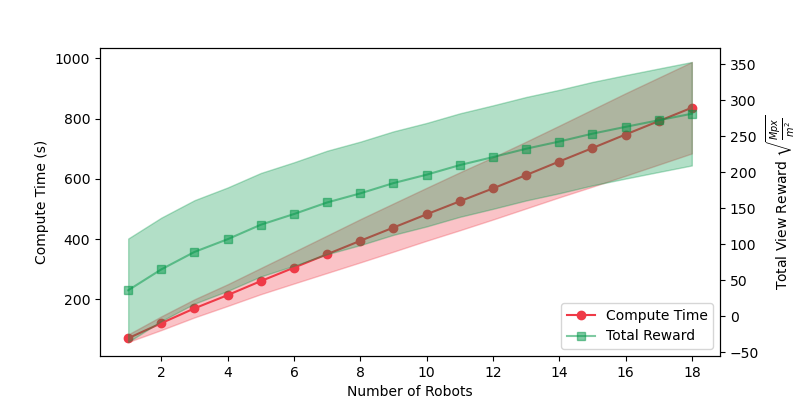}}
    \caption{Computation time and reward accumulated with scaling the number of robots in the \texttt{Large} test case. The means and standard deviations are computed across 10 unique start locations.}
    \label{fig:scale}
\end{figure}

Since $\viewreward$ is a square-root sum of pixel coverage over the
actors, we would expect scaling the number of robots to follow a similar
trend of diminishing returns.
In \cref{fig:scale}, we observe this trend with the $\texttt{Large}$ test case
for 1--18 robots.
These diminishing returns would
correspond---likely simultaneously---to increasing coverage over viewing angles
of the actors' faces and gradual (and ideally uniform) increase in pixel
densities.
Furthermore, scaling the number of robots produces nearly linear growth
in computation time even though our approach requires quadratic time
asymptotically (Sec.~\ref{sec:asymptotic_analysis}).
This is likely because the cost of inter-robot collision checks
is inconsequential compared to the cost of solving the single-robot MDPs.
Additionally, growth in planning time slows substantially following the first
robot---because robots share the same camera models we benefit significantly
from caching evaluation of camera views.

\section{Conclusion and Future Work}
In this work, we presented a novel system for multi-robot view planning based
on sequential greedy planning with an occlusion-aware objective.
Through evaluation in five different scenarios, we observe sequential planning
outperforming formation-based planning and specifically excelling in
obstacle-dense environments.
Additionally, we observe similar perception performance for sequential planning with and
without inter-robot collision constraints.
This demonstrates that sequential planning is able to find good solutions
when accounting for possible collisions between robots (despite no longer
satisfying requirements for bounded suboptimality).
In future work, we aim to extend this view planner to a 3D human pose
reconstruction task, and optimize our implementation to run at real-time rates.

\section*{Acknowledgment}
This work was supported by the National Science Foundation (NSF) under Grant No.
2024173.
Krishna was sponsored by the NSF REU program, as a part of the Robotics
Institute Summer Scholars (RISS) program.
Micah contributed to this work primarily while a postdoc at CMU.

\balance
{
\scriptsize
\bibliographystyle{IEEEtranN}
\bibliography{micah_abbreviations,references}

\begin{thebibliography}{38}
\providecommand{\natexlab}[1]{#1}
\providecommand{\url}[1]{#1}
\csname url@samestyle\endcsname
\providecommand{\newblock}{\relax}
\providecommand{\bibinfo}[2]{#2}
\providecommand{\BIBentrySTDinterwordspacing}{\spaceskip=0pt\relax}
\providecommand{\BIBentryALTinterwordstretchfactor}{4}
\providecommand{\BIBentryALTinterwordspacing}{\spaceskip=\fontdimen2\font plus
\BIBentryALTinterwordstretchfactor\fontdimen3\font minus \fontdimen4\font\relax}
\providecommand{\BIBforeignlanguage}[2]{{%
\expandafter\ifx\csname l@#1\endcsname\relax
\typeout{** WARNING: IEEEtranN.bst: No hyphenation pattern has been}%
\typeout{** loaded for the language `#1'. Using the pattern for}%
\typeout{** the default language instead.}%
\else
\language=\csname l@#1\endcsname
\fi
#2}}
\providecommand{\BIBdecl}{\relax}
\BIBdecl

\bibitem[Corah and Michael(2021{\natexlab{a}})]{corah_scalable_2021}
M.~Corah and N.~Michael, ``Scalable distributed planning for multi-robot, multi-target tracking,'' in \emph{Proc. of the {IEEE/RSJ} Intl. Conf. on Intell. Robots and Syst.}, Prague, Czech Republic, Sep. 2021.

\bibitem[Cai et~al.(2023)Cai, Schlotfeldt, Khosoussi, Atanasov, Pappas, and How]{cai2023energy}
X.~Cai, B.~Schlotfeldt, K.~Khosoussi, N.~Atanasov, G.~J. Pappas, and J.~P. How, ``Energy-aware, collision-free information gathering for heterogeneous robot teams,'' \emph{{IEEE} Trans. Robotics}, vol.~39, pp. 2585--2602, 2023.

\bibitem[Schlotfeldt et~al.(2021)Schlotfeldt, Tzoumas, and Pappas]{schlotfeldt2021tro}
B.~Schlotfeldt, V.~Tzoumas, and G.~J. Pappas, ``Resilient active information acquisition with teams of robots,'' \emph{{IEEE} Trans. Robotics}, vol.~38, no.~1, pp. 244--261, 2021.

\bibitem[Hung et~al.(2022)Hung, Hsu, and Cheng]{hung2022tcns}
H.-A. Hung, H.-H. Hsu, and T.-H. Cheng, ``Image-based multi-{UAV} tracking system in a cluttered environment,'' \emph{{IEEE} Transactions on Control of Network Systems}, vol.~9, no.~4, pp. 1863--1874, 2022.

\bibitem[Zhao et~al.(2019)Zhao, Wang, Wang, Cong, and Shen]{zhao2019aamas}
Y.~Zhao, X.~Wang, C.~Wang, Y.~Cong, and L.~Shen, ``Systemic design of distributed multi-{UAV} cooperative decision-making for multi-target tracking,'' \emph{Autonomous Agents and Multi-Agent Systems}, vol.~33, pp. 132--158, 2019.

\bibitem[Corah and Michael(2019)]{corah2019auro}
M.~Corah and N.~Michael, ``Distributed matroid-constrained submodular maximization for multi-robot exploration: theory and practice,'' \emph{Auton. Robots}, vol.~43, no.~2, pp. 485--501, 2019.

\bibitem[Bucker et~al.(2021)Bucker, Bonatti, and Scherer]{bucker_you_2021}
A.~Bucker, R.~Bonatti, and S.~Scherer, ``Do you see what {I} see? {C}oordinating multiple aerial cameras for robot cinematography,'' in \emph{Proc. of the {IEEE} Intl. Conf. on Robot. and Autom.}, Xi'an, China, May 2021.

\bibitem[Alc\'antara et~al.(2020)Alc\'antara, Capit\'an, Torres-Gonz\'alez, Cunha, and Ollero]{alcantara_autonomous_2020}
A.~Alc\'antara, J.~Capit\'an, A.~Torres-Gonz\'alez, R.~Cunha, and A.~Ollero, ``Autonomous execution of cinematographic shots with multiple drones,'' \emph{IEEE Access}, vol.~8, pp. 201\,300--201\,316, 2020.

\bibitem[Pueyo et~al.(2024)Pueyo, Dendarieta, Montijano, Murillo, and Schwager]{pueyo2024tro}
P.~Pueyo, J.~Dendarieta, E.~Montijano, A.~C. Murillo, and M.~Schwager, ``{CineMPC}: A fully autonomous drone cinematography system incorporating zoom, focus, pose, and scene composition,'' \emph{{IEEE} Trans. Robotics}, vol.~40, pp. 1740--1757, 2024.

\bibitem[N{\"a}geli et~al.(2017)N{\"a}geli, Alonso-Mora, Domahidi, Rus, and Hilliges]{nageli2017ral}
T.~N{\"a}geli, J.~Alonso-Mora, A.~Domahidi, D.~Rus, and O.~Hilliges, ``Real-time motion planning for aerial videography with dynamic obstacle avoidance and viewpoint optimization,'' \emph{{IEEE} Robot. Autom. Letters}, vol.~2, no.~3, pp. 1696--1703, 2017.

\bibitem[Ho et~al.(2021)Ho, Jong, Freeman, Rao, Bonatti, and Scherer]{ho_3d_2021}
C.~Ho, A.~Jong, H.~Freeman, R.~Rao, R.~Bonatti, and S.~Scherer, ``{3D} human reconstruction in the wild with collaborative aerial cameras,'' in \emph{Proc. of the {IEEE/RSJ} Intl. Conf. on Intell. Robots and Syst.}, Prague, Czech Republic, Sep. 2021.

\bibitem[Saini et~al.(2019)Saini, Price, Tallamraju, Enficiaud, Ludwig, Martinovic, Ahmad, and Black]{saini2019markerless}
N.~Saini, E.~Price, R.~Tallamraju, R.~Enficiaud, R.~Ludwig, I.~Martinovic, A.~Ahmad, and M.~J. Black, ``Markerless outdoor human motion capture using multiple autonomous micro aerial vehicles,'' in \emph{Proc. of the IEEE/CVF Intl. Conf. on Comp. Vis.}, Seoul, South Korea, 2019.

\bibitem[Cao et~al.(2022)Cao, Zhu, Yang, Xia, Choset, Oh, and Zhang]{cao2022icra}
C.~Cao, H.~Zhu, F.~Yang, Y.~Xia, H.~Choset, J.~Oh, and J.~Zhang, ``Autonomous exploration development environment and the planning algorithms,'' in \emph{Proc. of the {IEEE} Intl. Conf. on Robot. and Autom.}, Philadelphia, PA, May 2022.

\bibitem[Tallamraju et~al.(2019)Tallamraju, Price, Ludwig, Karlapalem, B{\"u}lthoff, Black, and Ahmad]{tallamraju2019ral}
R.~Tallamraju, E.~Price, R.~Ludwig, K.~Karlapalem, H.~H. B{\"u}lthoff, M.~J. Black, and A.~Ahmad, ``Active perception based formation control for multiple aerial vehicles,'' \emph{{IEEE} Robot. Autom. Letters}, vol.~4, no.~4, pp. 4491--4498, 2019.

\bibitem[Jiang and Isler(2023)]{jiang_onboard_2023}
\BIBentryALTinterwordspacing
Q.~Jiang and V.~Isler, ``Onboard view planning of a flying camera for high fidelity {3D} reconstruction of a moving actor,'' Jul. 2023. [Online]. Available: \url{http://arxiv.org/abs/2308.00134}
\BIBentrySTDinterwordspacing

\bibitem[Hughes et~al.(2024)Hughes, Martin, Corah, and Scherer]{hughes2024cdc}
S.~Hughes, R.~Martin, M.~Corah, and S.~Scherer, ``Multi-robot planning for filming groups of moving actors leveraging submodularity and pixel density,'' in \emph{Proc. of the {IEEE} Conf. on Decision and Control}, Milan, Italy, Dec. 2024, to appear.

\bibitem[noa()]{noauthor_drone_nodate}
\BIBentryALTinterwordspacing
``\BIBforeignlanguage{en-us}{Drone {That} {Follows} {You} - {Skydio} 2+ {\textbar} {Skydio}}.'' [Online]. Available: \url{https://www.skydio.com/skydio-2-plus/}
\BIBentrySTDinterwordspacing

\bibitem[Alc{\'a}ntara et~al.(2020)Alc{\'a}ntara, Capit{\'a}n, Torres-Gonz{\'a}lez, Cunha, and Ollero]{alcantara2020ieee}
A.~Alc{\'a}ntara, J.~Capit{\'a}n, A.~Torres-Gonz{\'a}lez, R.~Cunha, and A.~Ollero, ``Autonomous execution of cinematographic shots with multiple drones,'' \emph{IEEE Access}, vol.~8, pp. 201\,300--201\,316, 2020.

\bibitem[Mademlis et~al.(2023)Mademlis, Torres-Gonz{\'a}lez, Capit{\'a}n, Montagnuolo, Messina, Negro, Le~Barz, Gon{\c{c}}alves, Cunha, Guerreiro, et~al.]{mademlis2023}
I.~Mademlis, A.~Torres-Gonz{\'a}lez, J.~Capit{\'a}n, M.~Montagnuolo, A.~Messina, F.~Negro, C.~Le~Barz, T.~Gon{\c{c}}alves, R.~Cunha, B.~Guerreiro \emph{et~al.}, ``A multiple-{UAV} architecture for autonomous media production,'' \emph{Multimedia Tools and Applications}, vol.~82, no.~2, pp. 1905--1934, 2023.

\bibitem[Caraballo et~al.(2020)Caraballo, Montes-Romero, D{\'\i}az-B{\'a}{\~n}ez, Capit{\'a}n, Torres-Gonz{\'a}lez, and Ollero]{caraballo2020iros}
L.-E. Caraballo, {\'A}.~Montes-Romero, J.-M. D{\'\i}az-B{\'a}{\~n}ez, J.~Capit{\'a}n, A.~Torres-Gonz{\'a}lez, and A.~Ollero, ``Autonomous planning for multiple aerial cinematographers,'' in \emph{Proc. of the {IEEE/RSJ} Intl. Conf. on Intell. Robots and Syst.}, Las Vegas, Nevada, Sep. 2020.

\bibitem[Ray et~al.(2021)Ray, Pierson, Zhu, Alonso-Mora, and Rus]{ray2021iros}
A.~Ray, A.~Pierson, H.~Zhu, J.~Alonso-Mora, and D.~Rus, ``Multi-robot task assignment for aerial tracking with viewpoint constraints,'' in \emph{Proc. of the {IEEE/RSJ} Intl. Conf. on Intell. Robots and Syst.}, Prague, Czech Republic, May 2021.

\bibitem[Pueyo et~al.(2023)Pueyo, Montijano, Murillo, and Schwager]{pueyo2023iros}
P.~Pueyo, E.~Montijano, A.~C. Murillo, and M.~Schwager, ``{CineTransfer}: Controlling a robot to imitate cinematographic style from a single example,'' in \emph{Proc. of the {IEEE/RSJ} Intl. Conf. on Intell. Robots and Syst.}, Detroit, Michigan, Oct. 2023.

\bibitem[Bonatti et~al.(2020)Bonatti, Wang, Ho, Ahuja, Gschwindt, Camci, Kayacan, Choudhury, and Scherer]{bonatti2020jfr}
R.~Bonatti, W.~Wang, C.~Ho, A.~Ahuja, M.~Gschwindt, E.~Camci, E.~Kayacan, S.~Choudhury, and S.~Scherer, ``Autonomous aerial cinematography in unstructured environments with learned artistic decision-making,'' \emph{J. Field Robot.}, vol.~37, no.~4, pp. 606--641, 2020.

\bibitem[Xu et~al.(2022)Xu, Shi, Tokekar, and Diaz-Mercado]{xu2022iros}
X.~Xu, G.~Shi, P.~Tokekar, and Y.~Diaz-Mercado, ``Interactive multi-robot aerial cinematography through hemispherical manifold coverage,'' in \emph{Proc. of the {IEEE/RSJ} Intl. Conf. on Intell. Robots and Syst.}, Kyoto, Japan, Oct. 2022.

\bibitem[Tallamraju et~al.(2020)Tallamraju, Saini, Bonetto, Pabst, Liu, Black, and Ahmad]{aircaprl}
R.~Tallamraju, N.~Saini, E.~Bonetto, M.~Pabst, Y.~Liu, M.~Black, and A.~Ahmad, ``{AirCapRL}: Autonomous aerial human motion capture using deep reinforcement learning,'' \emph{{IEEE} Robot. Autom. Letters}, vol.~5, no.~4, pp. 6678--6685, 2020.

\bibitem[Delmerico et~al.(2018)Delmerico, Isler, Sabzevari, and Scaramuzza]{delmerico2018auro}
J.~Delmerico, S.~Isler, R.~Sabzevari, and D.~Scaramuzza, ``A comparison of volumetric information gain metrics for active {3D} object reconstruction,'' \emph{Auton. Robots}, vol.~42, no.~2, pp. 197--208, 2018.

\bibitem[Roberts et~al.(2017)Roberts, Shah, Dey, Truong, Sinha, Kapoor, Hanrahan, and Joshi]{roberts2017iccv}
M.~Roberts, S.~Shah, D.~Dey, A.~Truong, S.~Sinha, A.~Kapoor, P.~Hanrahan, and N.~Joshi, ``Submodular trajectory optimization for aerial {3D} scanning,'' in \emph{Proc. of the IEEE/CVF Intl. Conf. on Comp. Vis.}, Venice, Italy, Oct. 2017, pp. 5334--5343.

\bibitem[Nemhauser et~al.(1978)Nemhauser, Wolsey, and Fisher]{nemhauser1978}
G.~L. Nemhauser, L.~A. Wolsey, and M.~L. Fisher, ``An analysis of approximations for maximizing submodular set functions-{I},'' \emph{Math. Program.}, vol.~14, no.~1, pp. 265--294, 1978.

\bibitem[Fisher et~al.(1978)Fisher, Nemhauser, and Wolsey]{fisher1978}
M.~L. Fisher, G.~L. Nemhauser, and L.~A. Wolsey, ``An analysis of approximations for maximizing submodular set functions-{II},'' \emph{Polyhedral Combinatorics}, vol.~8, pp. 73--87, 1978.

\bibitem[Singh et~al.(2009)Singh, Krause, Guestrin, and Kaiser]{singh_efficient_2009}
A.~Singh, A.~Krause, C.~Guestrin, and W.~J. Kaiser, ``Efficient informative sensing using multiple robots,'' \emph{J. Artif. Intell. Res.}, vol.~34, pp. 707--755, 2009.

\bibitem[Lauri et~al.(2020)Lauri, Pajarinen, Peters, and Frintrop]{lauri_multi-sensor_2020}
M.~Lauri, J.~Pajarinen, J.~Peters, and S.~Frintrop, ``Multi-sensor next-best-view planning as matroid-constrained submodular maximization,'' \emph{{IEEE} Robot. Autom. Letters}, vol.~5, no.~4, pp. 5323--5330, 2020.

\bibitem[McCammon et~al.(2021)McCammon, Marcon~dos Santos, Frantz, Welch, Best, Shearman, Nash, Barth, Adams, and Hollinger]{mccammon2021jfr}
S.~McCammon, G.~Marcon~dos Santos, M.~Frantz, T.~P. Welch, G.~Best, R.~K. Shearman, J.~D. Nash, J.~A. Barth, J.~A. Adams, and G.~A. Hollinger, ``Ocean front detection and tracking using a team of heterogeneous marine vehicles,'' \emph{J. Field Robot.}, vol.~38, no.~6, pp. 854--881, 2021.

\bibitem[Corah(2022)]{corah_performance_2022}
M.~Corah, ``On performance impacts of coordination via submodular maximization for multi-robot perception planning and the dynamics of target coverage and cinematography,'' in \emph{{RSS} Workshop on Envisioning an Infrastructure for Multi-robot and Collaborative Autonomy Testing and Evaluation}, 2022.

\bibitem[Corah and Michael(2021{\natexlab{b}})]{corah2021icra}
M.~Corah and N.~Michael, ``Volumetric objectives for multi-robot exploration of three-dimensional environments,'' in \emph{Proc. of the {IEEE} Intl. Conf. on Robot. and Autom.}, Xi'an, China, May 2021.

\bibitem[Schrijver(2003)]{schrijver_combinatorial_2003}
A.~Schrijver, \emph{Combinatorial optimization: polyhedra and efficiency}.\hskip 1em plus 0.5em minus 0.4em\relax Springer Science \& Business Media, 2003, vol.~24.

\bibitem[Bargiacchi et~al.(2020)Bargiacchi, Roijers, and Now\'{e}]{noauthor_svalorzenai-toolbox_nodate}
E.~Bargiacchi, D.~M. Roijers, and A.~Now\'{e}, ``{AI-Toolbox}: A {C++} library for reinforcement learning and planning (with {Python} bindings),'' \emph{Journal of Machine Learning Research}, vol.~21, no. 102, pp. 1--12, 2020.

\bibitem[Bonatti et~al.(2019)Bonatti, Ho, Wang, Choudhury, and Scherer]{bonatti2019iros}
R.~Bonatti, C.~Ho, W.~Wang, S.~Choudhury, and S.~Scherer, ``Towards a robust aerial cinematography platform: Localizing and tracking moving targets in unstructured environments,'' in \emph{Proc. of the {IEEE/RSJ} Intl. Conf. on Intell. Robots and Syst.}, Macau, China, Nov. 2019.

\bibitem[Bishop et~al.(2010)Bishop, Fidan, Anderson, Do{\u{g}}an{\c{c}}ay, and Pathirana]{bishop2010auto}
A.~N. Bishop, B.~Fidan, B.~D. Anderson, K.~Do{\u{g}}an{\c{c}}ay, and P.~N. Pathirana, ``Optimality analysis of sensor-target localization geometries,'' \emph{Automatica}, vol.~46, no.~3, pp. 479--492, 2010.

\end{thebibliography}
}
\end{document}